\documentclass[10pt,twocolumn,letterpaper]{article}

\usepackage{iccv}
\usepackage{times}
\usepackage{epsfig}
\usepackage{graphicx}
\usepackage{amsmath}
\usepackage{amssymb}
\usepackage{color}
\usepackage{caption}
\usepackage{subcaption}
\usepackage[pagebackref=true,breaklinks=true,letterpaper=true,colorlinks,bookmarks=false]{hyperref}

\iccvfinalcopy 

\def\iccvPaperID{16} 

\ificcvfinal\fi
\begin{document}

\title{Virtual Blood Vessels in Complex Background using Stereo X-ray Images}

\author{Qiuyu Chen\\
University of Washington\\
Seattle, WA, US\\    
{\tt\small qiuyuchen14@gmail.com}
\and
Ryoma Bise\\
Kyushu University\\
Fukuoka, Japan\\
{\tt\small bise@ait.kyushu-u.ac.jp}
\and
Lin Gu\\
National Institute of Informatics\\
Tokyo, Japan\\
{\tt\small lingu.edu@gmail.com}
\and
Yinqiang Zheng\\
National Institute of Informatics\\
Tokyo, Japan\\
{\tt\small yqzheng@nii.ac.jp}
\and
Imari Sato\\
National Institute of Informatics\\
Tokyo, Japan\\
{\tt\small imarik@nii.ac.jp}
\and
Jenq-Neng Hwang\\
University of Washington\\
Seattle, WA, US\\
{\tt\small hwang@uw.edu}
\and
Nobuaki Imanishi\\
Keio University\\
Tokyo, Japan\\
{\tt\small nimanmed@z5.keio.jp}
\and
Sadakazu Aiso \\
Keio University\\
Tokyo, Japan\\
{\tt\small aiso@a3.keio.jp}
}

\maketitle

\begin{abstract}
We propose a fully automatic system to reconstruct and visualize 3D blood vessels in Augmented Reality (AR) system from stereo X-ray images with bones and body fat. Currently, typical 3D imaging technologies are expensive and carrying the risk of irradiation exposure. To reduce the potential harm, we only need to take two X-ray images before visualizing the vessels. Our system can effectively reconstruct and visualize vessels in following steps. We first conduct initial segmentation using Markov Random Field and then refine segmentation in an entropy based post-process. We parse the segmented vessels by extracting their centerlines and generating trees. We propose a coarse-to-fine scheme for stereo matching, including initial matching using affine transform and dense matching using Hungarian algorithm guided by Gaussian regression. Finally, we render and visualize the reconstructed model in a HoloLens based AR system, which can essentially change the way of visualizing medical data. We have evaluated its performance by using synthetic and real stereo X-ray images, and achieved satisfactory quantitative and qualitative results. 
\end{abstract}

\section{Introduction}

Over years, advanced Augmented Reality (AR) and Virtual Reality (VR) techniques have been utilized to train surgeons, prepare operation procedure, improve the diagnosis, provide intraoperative data\cite{Alaraj11} and even remote examination \cite{Ruffaldi2015}. The VR system provides stereoscopic 3D visualization of vascular  \cite{Alaraj11}, micro calcifications  \cite{Douglas16}, liver \cite{Pessaux2015} \text{etc}.  which could be rotated, translated and zoomed by the user in an immerse environment. When detecting the offending vessels in neurovascular compression syndrome, AR/VR system has offered significantly improvement over traditional 2D images \cite{Kin09}.

\begin{figure*}[tp]
	\begin{subfigure}{.3\textwidth}
		\centering
		\includegraphics[width=5.5cm,height=5cm]{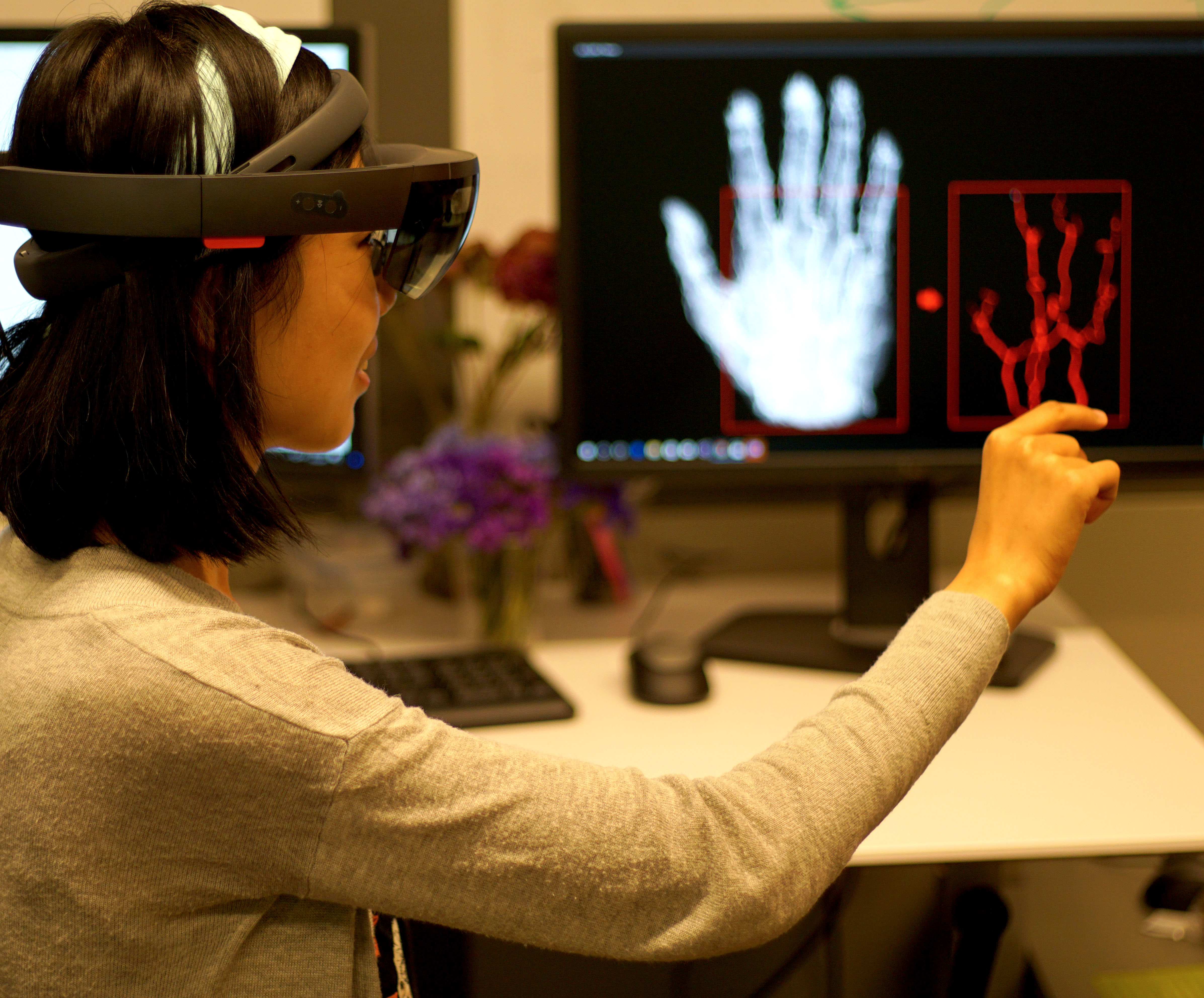}
		\caption{HoloLens system}
	\end{subfigure}\hfill
	\begin{subfigure}{.3\textwidth}
		\centering
		\includegraphics[width=5.5cm,height=5cm]{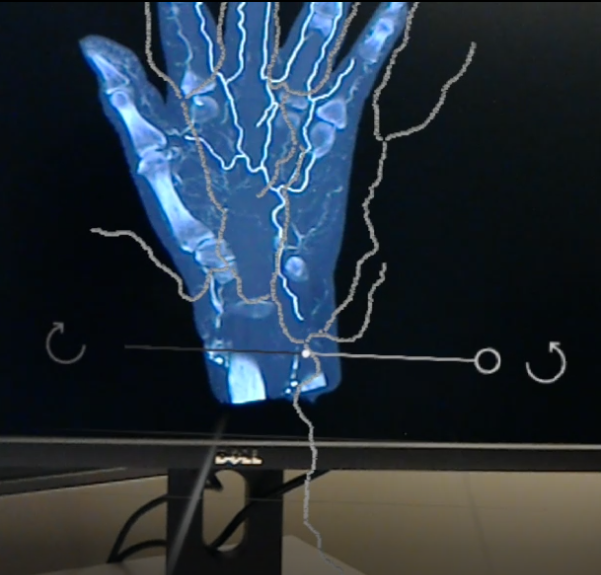}
		\caption{Virtual blood vessel from z-axis}
	\end{subfigure}\hfill
	\begin{subfigure}{.3\textwidth}
		\centering
		\includegraphics[width=5.5cm,height=5cm]{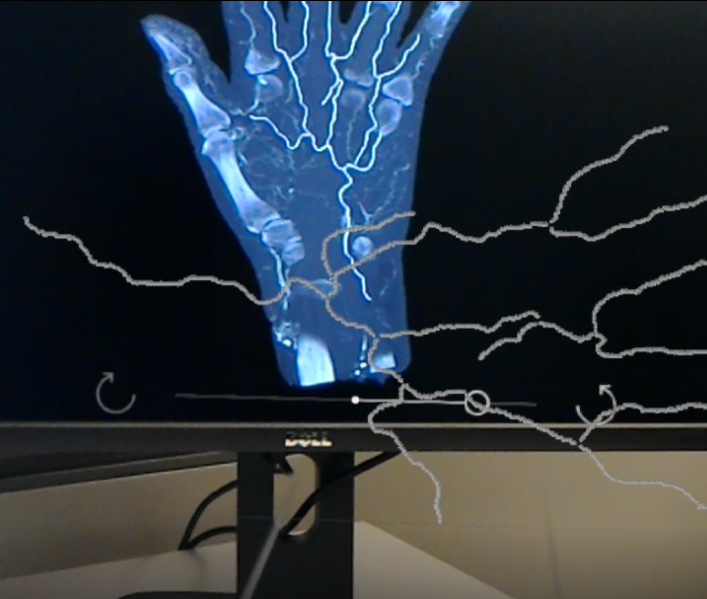}
		\caption{Virtual blood vessel from y-axis}
	\end{subfigure}
	\caption{Visualized 3D blood vessels in our Augmented Reality system}
	\label{fig:AR_system}
\end{figure*}

\begin{figure*}[t!h]
\begin{center}
\fbox{\includegraphics[height=8.5cm,width=17.5cm]{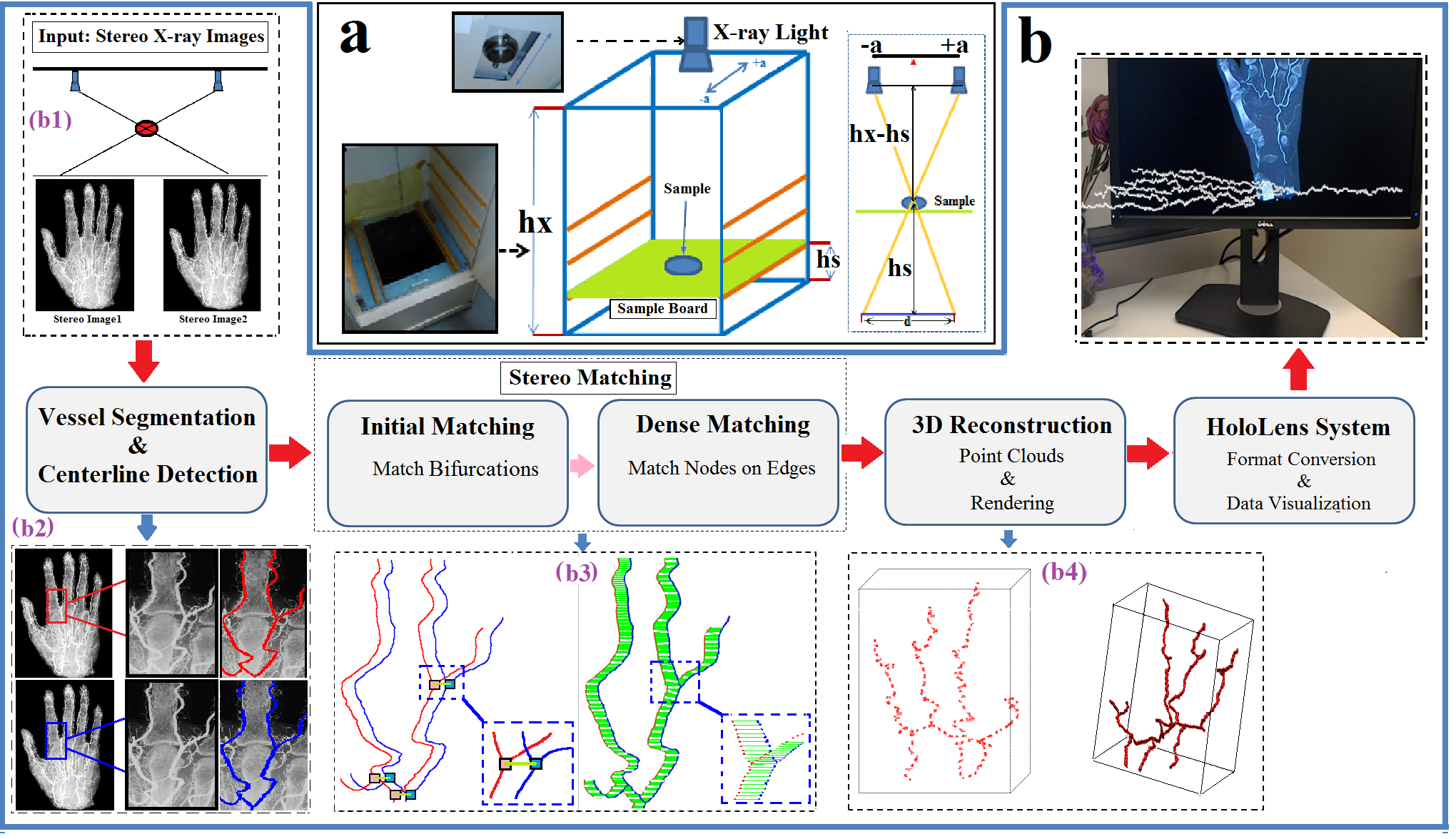}}	
\end{center}
   \caption{a. Equipment for stereo X-ray images. b. System workflow with examples for each steps}
\label{fig:workflow} 
\end{figure*}
As shown in Fig.\ref{fig:AR_system}, we propose a novel hololens\footnote{https://www.microsoft.com/en-us/hololens} based AR system that reconstructs and visualizes 3D structure of blood vessels to facilitate the medical research. Visualization of vessels would allow practitioners to establish correct diagnosis and further reduce the threat of diseases like cardiovascular and cancer  \cite{Moussawi2015,Qin2016}. To reconstruct blood vessels in 3D, current methods usually use 3D volume data captured by computed tomography angiographic (CTA) or magnetic resonance imaging (MRI). Since CTA typically requires multiple X-ray scans of the target area, whose radiation exposure is much larger than that of single X-ray imaging. Methods like biplane angiography usually need pre-defined models or use adopt point-based reconstruction methods in the absence of complex background  \cite{Chen97,Gu_2015_ICCV}. In recent years, simpler imaging systems have been developed. For example, Hoshino et al  \cite{hoshino2011} proposed an X-ray stereo imaging system that can record simultaneously X-ray stereo images using two beam paths. Though this system can easily take X-ray stereo images, it further requires a stereo matching method that reconstruct 3D vascular structures from two images. However, vessel detection and matching in the presence of complex background itself is extremely challenging. 

We illustrate our pipeline in Fig.\ref{fig:workflow}. The X-ray imaging system, based on the Softex C\footnote{http://www.softex-kk.co.jp/} system, used in this work is shown in Fig.\ref{fig:workflow}. With the planner detector and object fixed, we capture a pair of stereo images by moving the X-ray tube from position \textbf{-a} and \textbf{+a}.  From raw stereo X-ray images, we automatically reconstruct the 3D vessels in the following four step: 1. Given a pair of X-ray images with complex background of vessels (Fig.\ref{fig:workflow}.b1), we first apply vessel segmentation to extract centerlines (red and blue label in Fig.\ref{fig:workflow}.b2). 2. Stereo matching is then conducted to match a pair of vessel centerlines (Fig.\ref{fig:workflow}.b3). Specifically, this step can be divided into initial matching and dense matching. Initial matching (Fig.\ref{fig:workflow}.b3-left) only matches bifurcations using affine transform generated by SIFT  \cite{C8} on raw images. 3. Dense matching uses the Hungarian algorithm with Gaussian regression  \cite{SerradellPami15} to assign each pair of correspondence along every edge linking between two matched bifurcations, as shown as green lines in Fig.\ref{fig:workflow}.b3-Right. 4. Using the correspondence disparity in two stereo images, we can generate 3D point clouds and further render 3D vessel models as shown in Fig.\ref{fig:workflow}.b4. Finally we establish a HoloLens based AR system (Fig.\ref{fig:AR_system}.a) which allows user to view the reconstructed 3D vessels in an interactive immerse virtual environments. Users can rotate the model and check the vessel structure from different views (Fig.\ref{fig:AR_system}.bc)


In this paper, we have following four major contributions: 
\begin{enumerate}
	\item  We develop an effective segmentation method to detect vessels under complex background mixed with bones and body fat. 
	\item  By properly parsing the 2D vascular structure, we propose a new stereo matching method that outperforms the state-of-the-art.
	\item Our stereo based system could obtain 3D vascular structures, with much less health risk due to radiation exposure, shorter imaging time and significantly simplified procedures.
	\item  The reconstructed 3D models could be visualized in AR system. 
\end{enumerate}

\begin{figure*}
\begin{center}
\fbox{\includegraphics[height=6cm,width=17.5cm]{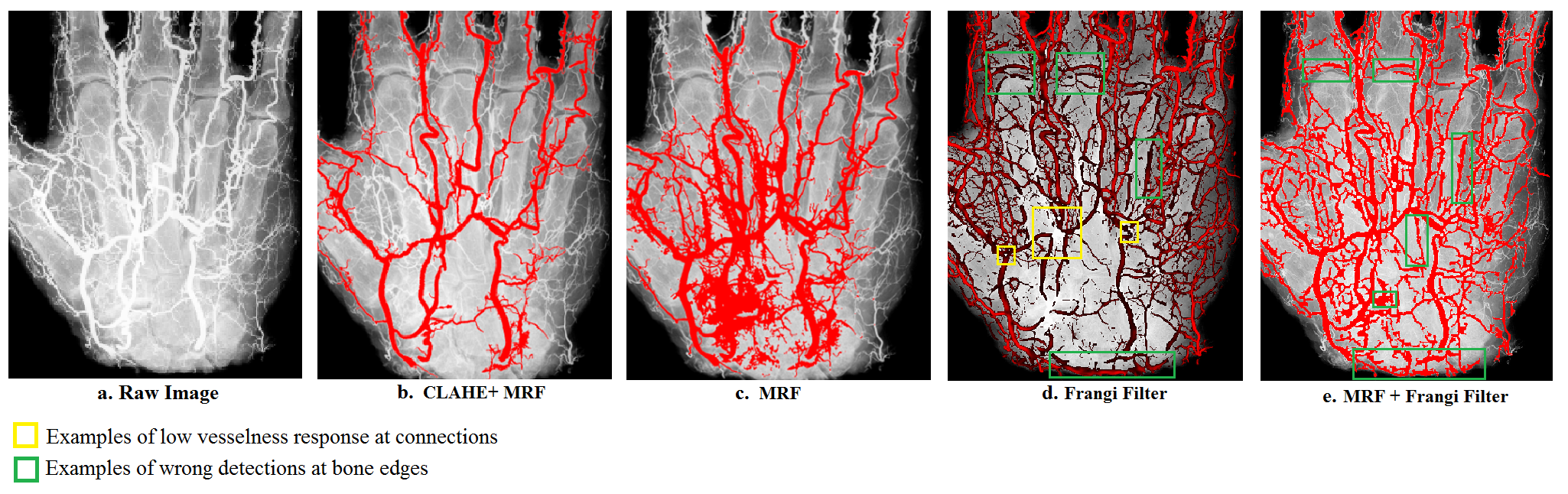}}	
\end{center}
   \caption{Segmentation Evaluation using Different Methods. a: raw image. b. Initial Segmentation using CLAHE and MRF. c: MRF. d:Frangi Filter e:MRF multi-label Optimization.}
\end{figure*}

\begin{figure*}
\begin{center}
\fbox{\includegraphics[height=4.5cm,width=17.5cm]{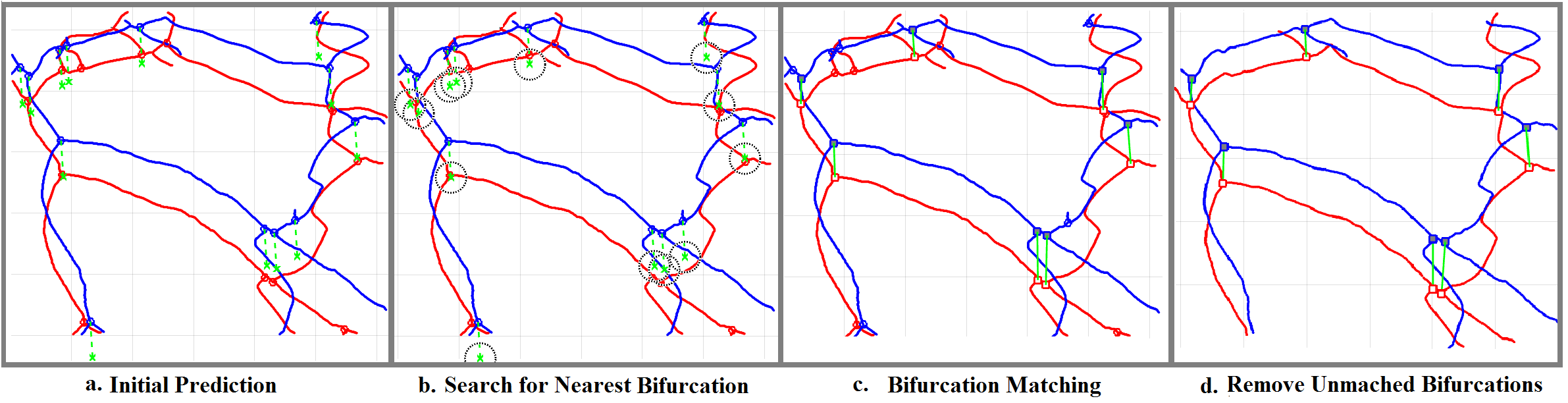}}	
\end{center}
   \caption{Stereo Matching Process. a. Initial prediction of corresponding bifurcation nodes using affine transform. b. Search for bifurcation nodes with minimum euclidean distance on targeted tree. Initial alignment.Blue dots represent warped image while red dots represent target image. c. Junction nodes matching. d. Remove branches with unmatched bifurcations}
   \label{fig:matching}
\end{figure*}

\section{Robust Segmentation of Blood Vessels under Complex Background}
\label{sec:segmentation}

Segmentation is an essential step to extract the vessel from an input raw image. We apply a Contrast Limited Adapted Histogram Equalization (CLAHE) \cite{Zuiderveld:1994} method, then we apply  Markov Random Field \cite{Kato2001} to divide the enhanced image into several segmentations with labels. Based on the observation from raw images that vessel regions contribute slightly higher intensity, we look for pixels with high intensity and associate them with segmentation labels. The majority of labels in this set indicates the label of segmentation for vessel regions.

Blood vessel segmentation has been studied for many years. One of the widely used vessel segmentation methods is proposed by Frangi \textit{et al.} \cite{Frangi1998} that measures the vesselness by computing the eigenvalues of Hessian at a certain scale. The possibility of tubular structure is the maximum vesselness response across several selected scales which could be automatically determined by  \cite{MirzaalianCVPR10} using Markov random field (MRF). However, existing Frangi filter \cite{Frangi1998} and its combination with MRF  \cite{MirzaalianCVPR10} suffer from two limitations, especially in current application: 1. Low vesselness response at vessel bifurcations where shapes are complex as illustrated in the yellow box in Fig.3.d. 2. High vesselness response at background bone edge due to its high intensity illustrated in the green box in Fig.3.d and Fig.3.e.   

Since these two types of errors would seriously spoil the succeeding stereo matching and 3D reconstruction steps described in Sec\ref{sec:3D_reconstruction}, our segmentation method can reliably enhance the vessel from complex background such as high intensity bones. Adaptive Histogram Equalization (AHE) enhances the contrast between bones and vessels by amplifying the vicinity of each pixel by its neighborhood cumulative distribution function (CDF). Since some noise may also be over amplified, CLAHE further limits the CDF, to help equally redistribute among all histogram bins.

To further remove the noise such as bone edges and tiny vessels, we conduct series of entropy based morphological post-process. The vessel should show relative lower entropy because it is usually smoother than the bone. Therefore, we compute the local entropy on the raw image masked by initial segmentation and remove any pixel that has high entropy. This would also help us to remove thin vessels since vessel edges also have high entropy. The minimum thickness of the reconstructed vessel depends on the segmentation and smoothing process. Then we only extract largest segmentation and fill holes through a morphological reconstruction algorithm in Matlab toolbox \cite{Laporte2003}. Finally, fine vessel centerlines are generated using thinning rule in  \cite{Lam92}. A graphic tree for each vessel centerline, which is represented by bifurcation nodes and edge nodes, is generated by a graph-based tracing method proposed by \cite{DeLiChe:Bmcbioinfo14}.

\section{3D Reconstruction of Blood Vessels}
\label{sec:3D_reconstruction}
In order to get 3D model of the blood vessel, we first do stereo matching in a coarse-to-fine scheme. In the initial matching, we match bifurcations using affine transform and then obtain dense correspondences using Hungarian Algorithm. We are able to reconstruct the correspondences in 3D point clouds and then render them into 3D models.  
\subsection{Stereo Matching}
\label{sec:matching}
Stereo Matching involves two steps: Initial Matching for bifurcations and Dense Matching for all the other nodes. Let \textbf{P} and  \textbf{Q} represent bifurcation sets on warped image and target image, $p\textsubscript{i}$ and $q\textsubscript{i}$ represent bifurcation i on warped and target image. 

\begin{figure*}[htb]
\begin{center}
		\includegraphics[width=\linewidth]{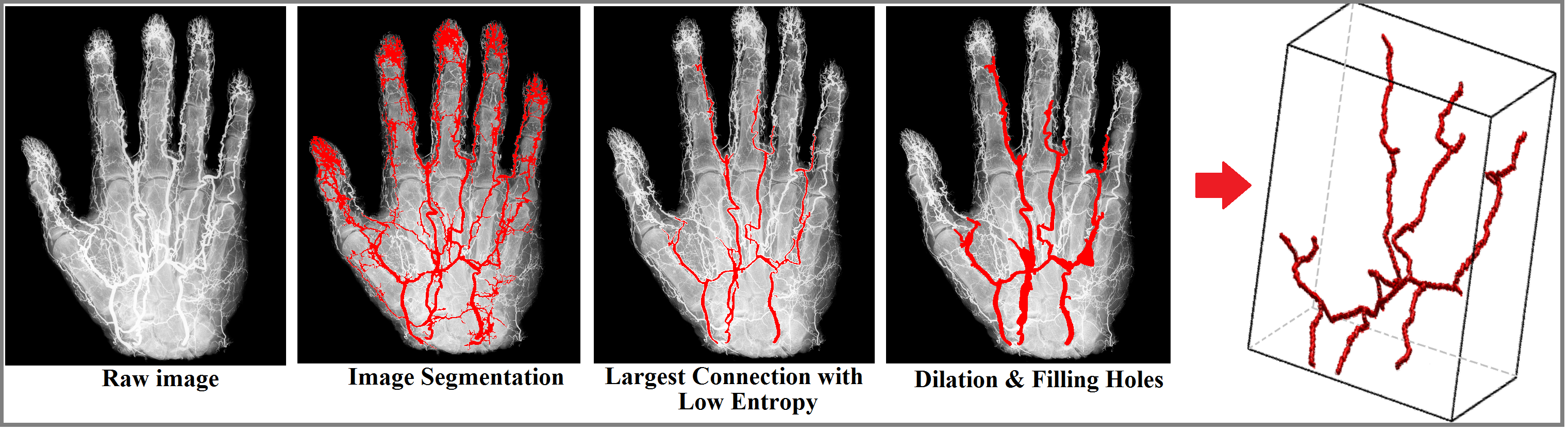}
	\end{center}
		\caption{3D reconstruction from stereo X-ray images of human hand. We first use CLAHE based MRF to segment blood vessels, and then extract largest connected segment and finally do standard morphological dilation and filling holes procedures.}
		\label{fig:real_exp}
\end{figure*}
\subsubsection{Initial Matching:}
In order to predict the correspondences on target image, we first calculate \(3\times3\) affine transform using SIFT  \cite{C8,C15} on the raw image. At each bifurcation $(p_x, p_y)$ on warped image, we predict its correspondence $(q\prime_x, q\prime_y)$ (Fig4.a). We use homogeneous coordinates to represent affine transform in a \(3\times3\) matrix and homogeneous vector for each bifurcation pixel:  
\begin{equation}
\begin{bmatrix}
q\prime_x\\
q\prime_y\\
1
\end{bmatrix}
=
\begin{bmatrix}
\textbf{R} & \textbf{T} \\
\textbf{0} & 1 
\end{bmatrix}
\begin{bmatrix}
p_x\\
p_y\\
1
\end{bmatrix}
\end{equation}
where \textbf{R} is a \(2\times2\) rotation matrix while \textbf{T} is a \(2\times1\) translation vector. For each prediction, we then search for it's nearest bifurcation in \textbf{Q} within a range r (Fig.\ref{fig:matching}b,c). We then remove edges between each unmatched bifurcation and it's connected terminal nodes (Fig.\ref{fig:matching}.d) and finally update trees for two images with $N_J$ matched bifurcations. 
\subsubsection{Dense Matching:}

The matched bifurcations suggest an initial mapping between two trees. A fine alignment approach using Hungarian algorithm guided by Gaussian regression was proposed by  \cite{SerradellPami15} to establish new matches between the edge points of the two paths. If we define the tree of warped image and targeted image as  \textbf{X\textsuperscript{A}} and \textbf{X\textsuperscript{B}}, then the correspondence set can be written as: $\pi$ = \{$x^A_i \leftrightarrow x^B_i$\}\textsubscript{$1\leq i \leq N$}, N is the number of correspondence. Use Gaussian non-linear regression, we can predict the location of new correspondence of points in \textbf{X\textsuperscript{A}} by: 
\begin{align*}
m_\pi(x^B)&=\textbf{k}^T\textbf{C}^{-1}_\pi \textbf{X}^A_\pi, \\ \sigma^2(x^B)&=k(x^B,x^B)+\beta^{-1}-\textbf{k}^T\textbf{C}^{-1}_\pi \textbf{k}
\end{align*}
where $\beta$ is the measurement noise variance, $k$ is a non-linear kernel function defined by Eq.2 and $\textbf{C}_\pi$ is \(N\times N\) matrix with element $C_{i,j}=k(x^B_i,x^B_j)+\beta ^{-1}\delta_{i,j}$, $\textbf{k}$ is the vector $[k(x^B_1,x^B),...,k(x^B_N,x^B)]^T$, with the kernel defined as
\begin{equation}
k(x_i,x_j)=\theta _0 + \theta _1x^T_ix_j+\theta _2\exp{\{-\frac{\theta _3}{2}\|x_i-x_j\|^2\}}
\end{equation}
 which can implicitly define a mapping function consist of both non-linear and affine transform in biomedical warping. Here $\theta _0$, $\theta _1$, $\theta _2$ and $\theta _3$ are hyperparameters, which are used for adjusting deformation weights.

We initialize the mapping using $N_J$ sets of correspondences from initial matching: $\pi$ = \{$p_i \leftrightarrow q_i$\}\textsubscript{$1\leq i \leq N_J$}. Following steps of fine alignment in \cite{SerradellPami15}, for each pair of edges connected by matched bifurcation, we find new correspondences. To evaluate dense mapping, we calculate the Euclidean distance between the point and its predicted correspondence. We associate it with Hungarian matrix and by finding minimum total cost and finally obtain dense matching \cite{SerradellPami15}.

\subsection{Reconstruction}
After we get correspondences: $\pi$ = \{$x^A_i \leftrightarrow x^B_i$\}\textsubscript{$1\leq i \leq N$}, we estimate each point in world coordinate $\textbf{X}^A_w$ and $\textbf{X}^B_w$ using intrinsic matrix of the camera \textbf{M}, provided by the manufacturer:  
\begin{equation}
\textbf{X}_w=
\begin{bmatrix}
x_w\\
y_w\\
1
\end{bmatrix}
=
\textbf{M}^{-1}
\begin{bmatrix}
x_i\\
y_i\\
1
\end{bmatrix}
\end{equation}
As shown in Fig.\ref{fig:workflow}a, we assume X-ray light rail is parallel to image plane. If we define hx is the height of X-ray source, hs is the height of the samples, and depth $z$ is hx-hs, we can get Eq.4 based on similar triangles theorems: 
\begin{equation} 
\dfrac{2a}{d}=\dfrac{hx-hs}{hs}=\dfrac{z}{hx-z}
\end{equation}
where d is the distance between two correspondences. Since two X-ray lights are symmetric to the sample position, finally we estimate a 3D point $[x, y, z]^T$ as: 
\begin{equation}
\begin{bmatrix}
x\\
y\\
z
\end{bmatrix}
=\dfrac{1}{2}\times (\textbf{X}^A_w + \textbf{X}^B_w ) \\
+
\begin{bmatrix}
0\\
0\\
2a\times\dfrac{hx}{d+2a}-1
\end{bmatrix}
\end{equation}
We further smooth the reconstruction by averaging depth within a step size along each branch in the 3D tree. We approximate the blood vessels as a series of small cylinders and thus described by SWC format, a widely used format to define the neuron and vessel structure. At each node of the vessel model, the radius vessel can be calculated by searching the first non-vessel pixel along the normal vector in the segmentation. Finally, 3D rendering is done by approximating a cylinder with estimated radius at each connection between two adjacent nodes in the tree. This step is done in Matlab Trees Toolbox.  

\begin{figure*}
\begin{center}
	\fbox{\includegraphics[height=6.5cm,width=17cm]{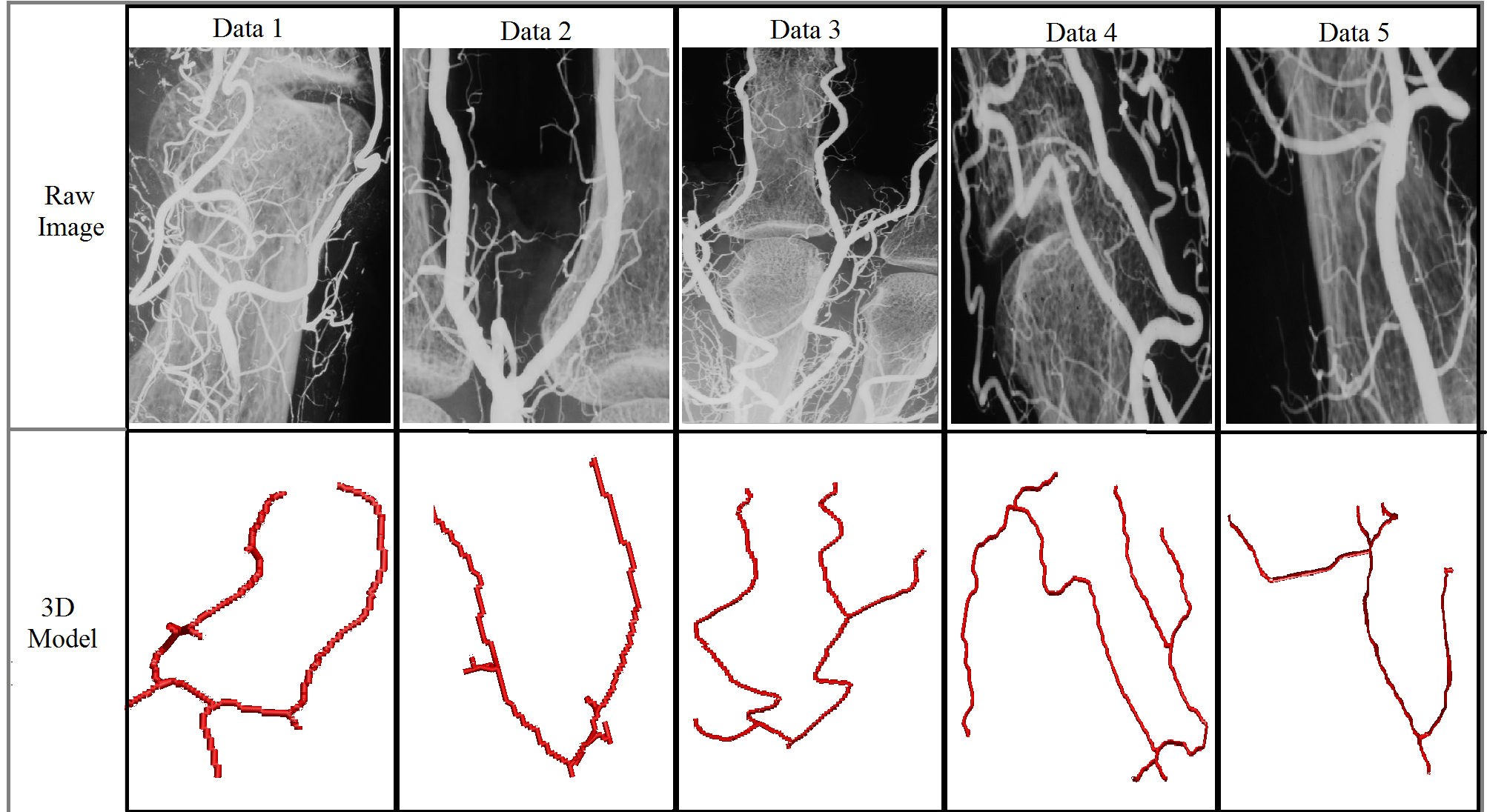}}
	\end{center}
	\caption{Experiments on 5 X-ray stereo images with bones. The first line shows 5 raw images, the second line shows their corresponding 3D rendering results using our method}
	\label{fig:rea_exp_detail}
\end{figure*}
\begin{figure*}
\begin{center}
\fbox{\includegraphics[height=6cm,width=17.5cm]{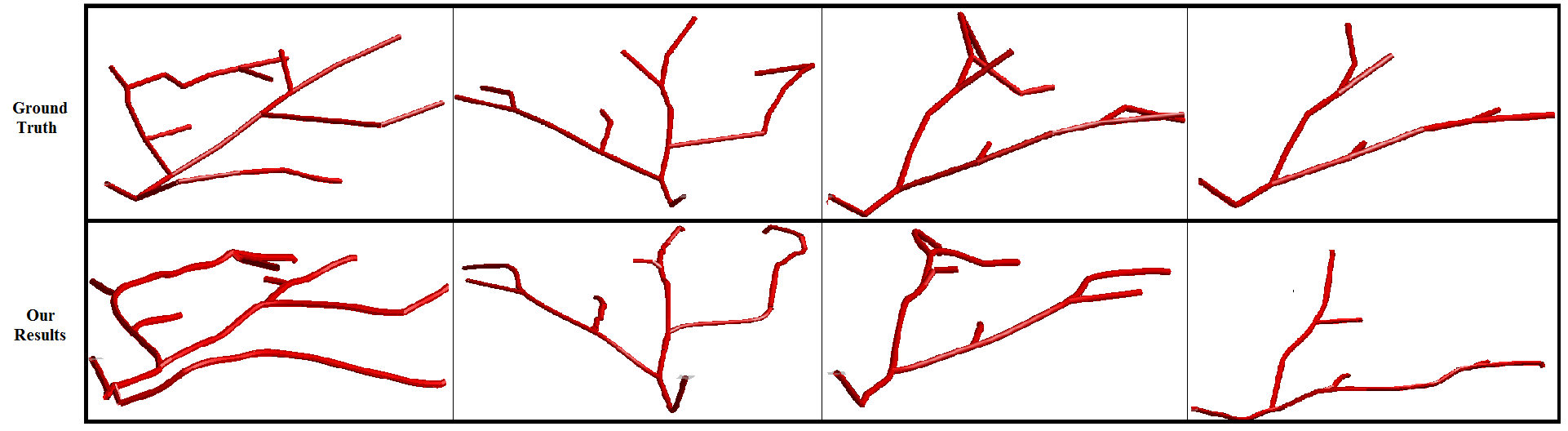}}	
\end{center}
   \caption{Synthetic Experiments: First row: Ground truth generated by Matlab Tree Toolbox. Second row: 3D reconstruction using our system.}
\label{fig:synthetic}   
\end{figure*}

\begin{figure*}[t]
\begin{center}
\fbox{\includegraphics[height=8.5cm,width=16.5cm]{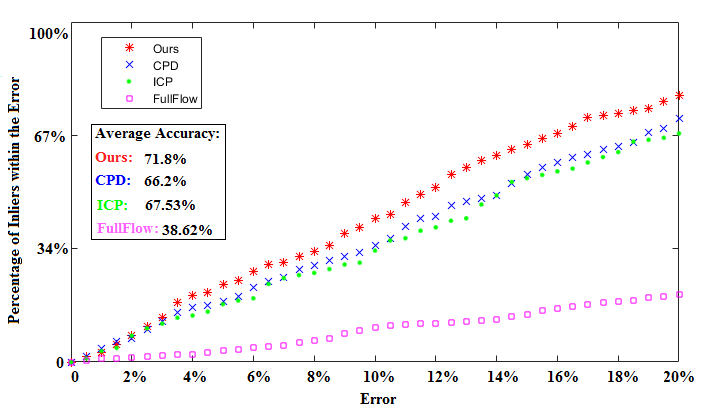}}	
\end{center}
   \caption{Accuracy Evaluations for our method, CPD and ICP and fullflow. }
\label{fig:eval}
\end{figure*}

\section{Visualizing 3D Models using Augmented Reality System}
\label{sec:VR}
\subsection{HoloLens System}

Augmented reality headsets, such as Microsoft HoloLens  \cite{Hololens}, move data visualization from 2D screen to 3D hologram-like interface. It can project a mixed-reality overlay on real world, and may lead to a large impact in medical applications. 
To facilitate the diagnosis, we import the reconstructed 3D vessel model in the virtual reality system where doctors could view the vessel of interest in any angle.

In order to visualize vessel, after computing the reconstructed data from stereo X-ray images, we would upload the 3D vessel model into the cloud and import into 3D viewer beta in HoloLens. Specifically, since the HoloLens system visualizes the target by rendering the 3D model of fbx format in front of the user, we have to build the model in DXF file format before converting into fbx 2013 format. Finally, as shown in Fig.\ref{fig:AR_system},  users could rotate, zoom, and move the reconstructed vessels from raw images to better assist doctors' work.  

\section{Experiment}
\label{sec:Exp}

\subsection{3D Vessel Reconstruction and Visualization from X-ray Stereo Images}
To begin with, we demonstrates the results of our system on real world data collected by our X-ray imaging equipment (Fig.\ref{fig:workflow}.a) for clinic practice. In Fig.\ref{fig:real_exp}, we can observe that the raw input images contain a huge portion of bone in the background. The vessels also assume little surface texture, which would frustrate existing standard stereo matching methods, either sparse or dense. Our vessel segmentation method successfully localizes main vessels despite the presence of complex background. We first apply CLAHE on two stereo X-ray images and use MRF for multi-labeled segmentation. We search for pixels with highest intensity from raw images and associate them with a MRF label. Then we only extract pixels with that label and remove small segments by extracting largest connected pixels. Finally, we do dilation and fill holes in order to get tree structures. Our stereo matching and reconstruction method captures the disparity properly, and the reconstructed 3D vessels look reasonable and satisfactory. We also show our reconstruction results in five small regions from X-ray images in Fig.\ref{fig:rea_exp_detail}. Our system works well for large and less occluded vessels. However, due to the restriction of view angles, it may fail for heavily occluded vessels. In our dataset, the minimum thickness of vessels is about 10 pixels.

\subsection{Quantitative Evaluation of Stereo Matching}

To evaluate the accuracy of our novel stereo matching algorithm, we compare our results with CT scan. A z-slice image ($512 \time 512$) was reconstructed from the 2496 signals captured by 64 array sensor (GE Discovery CT750 HD). We simulate a pair of stereo images by projecting the 3D volume data into two planes. Due to high occlusion of bones and low resolution, we manually segment vessels and test accuracy of method in section 4.2. We associate our points with nearest human-labeled points on 2D CT image and evaluate each depth accuracy. We only care about the relative depth in this test so both CT and our results are normalized. We compare our results with state-of-art stereo matching methods: Fullflow  \cite{Chen16CVPR} and the other non-rigid point matching methods including Coherent Point Drift (CPD)  \cite{C51} and Iterative Closest Point (ICP)  \cite{bergstrom2014}. At each point, we define error as $e\%=\frac{|Depth-GT|}{GT}$, where depth is the calculated depth result $z$ while ground truth is depth obtained by CTA. For each method, we calculate the average error of all the points: $\sqrt{\frac{\sum\limits_{i=1}^n e^2_i}{n}}$, n is the number of points. Fig.\ref{fig:eval} is the error histogram, which is generated by counting how many points have errors within this value. We achieve 72.6\% points with less than 30\% error, with an average accuracy of 71.8\%. 

\subsection{Synthetic Experiment}
We also design four synthetic vessel skeletons and render them as ground truth in Matlab Trees Toolbox as shown in Fig.\ref{fig:synthetic}. We define locations of nodes and width of vessels and connect them by cylinders. In order to validate the robustness of our reconstruction system, we specifically introduce several corners and large angle in the synthetic vessels which is challenging for stereo vision. We simulate the experiment based on the principle of our equipment and project the 3D skeleton model into two stereo images. Because the projected stereo skeleton might be sparse, we further generate fine centerlines using morphological process. Using methods described in Section\ref{sec:3D_reconstruction}, we generate corresponding 3D models and compare them with ground truth qualitatively.   

\section{Conclusion}
\label{sec:conclusion}
In this paper, we propose an Augmented Reality (AR) system to reconstruct and visualize 3D blood vessels from stereo X-ray image under complex backgrounds. Our system could obtain 3D vascular structures, with much less health risk due to radiation exposure, shorter imaging time and significantly simplified procedures. Comparing with state-of-the-art, our novel stereo matching algorithm achieves a higher accuracy on both real and synthetic data. 

\section*{Acknowledgement}

This work was funded by ImPACT Program of Council for Science, Technology and Innovation (Cabinet Office, Government of Japan).

{\small
\bibliographystyle{ieee}
\bibliography{egbib}
}

\end{document}